\documentclass[preprint,secnumarabic,amssymb, nobibnotes, aps, pra]{revtex4-1}
\usepackage{graphicx}
\usepackage{amsmath}

\begin{document}

\title{Exploiting non-i.i.d. data towards more robust machine learning algorithms}%

\author{Wim Casteels}%
\email{wim.casteels@uantwerpen.be}
\author{Peter Hellinckx}%
\affiliation{University of Antwerp - imec, IDLab - Faculty of Applied Engineering,
Sint-Pietersvliet 7, 2000 Antwerp, Belgium}

\begin{abstract}
In the field of machine learning there is a growing interest towards more robust and generalizable algorithms. This is for example important to bridge the gap between the environment in which the training data was collected and the environment where the algorithm is deployed. Machine learning algorithms have increasingly been shown to excel in finding patterns and correlations from data. Determining the consistency of these patterns and for example the distinction between causal correlations and nonsensical spurious relations has proven to be much more difficult. In this paper a regularization scheme is introduced that prefers universal causal correlations. This approach is based on 1) the robustness of causal correlations and 2) the data not being independently and identically distribute (i.i.d.). The scheme is demonstrated with a classification task by clustering the (non-i.i.d.) training set in subpopulations. A non-i.i.d. regularization term is then introduced that penalizes weights that are not invariant over these clusters. The resulting algorithm favours correlations that are universal over the subpopulations and indeed a better performance is obtained on an out-of-distribution test set with respect to a more conventional $l_2$-regularization.
\end{abstract}

\maketitle

\section{Introduction}
A cornerstone of learning theory is that the data is assumed to be independent and identically distributed (i.i.d.). This assumption is pivotal to draw conclusions about the generalization error within the Probably Approximately Correct (PAC) mathematically framework \cite{valiant1984theory}. However in practice the i.i.d. assumption is in many cases not strictly valid which makes makes it difficult to estimate the general performance of an ML algorithm. A major challenge in machine learning is to make the algorithms more robust such that their performance does not deteriorate when confronted with changes in the data distribution.

Recently a non-i.i.d. (ni) index was introduced in Ref. \cite{he2020towards} that quantifies the degree of distribution shift between training data and test data (also generally known as covariate shift). They use this index to quantify the violation of i.i.d. for a specific data set by randomly splitting the dataset in a training and a test set and calculating the ni index. For example for the celebrated ImageNet dataset they reveal that the ni index is larger than zero for all considered classes. We will show in this paper how such a violation of i.i.d. can be leveraged towards more robust algorithms.

Closely related to the general goal towards more robust algorithms is the topic of causality which has also been attracting a lot of interest in recent years (see for example Refs. \cite{peters2017elements, rohrer2018thinking, hassani2018big, pearl2019seven, scholkopf2019causality, 10.1145/3397269}). Many machine learning algorithms excel in finding patterns or correlations in a dataset but they are mostly oblivious to the origin or cohesion of these patterns. On the other hand it is well known that the likelihood that some nonsensical spurious correlation exists increases as more datafields are considered which becomes increasingly applicable in the current big data era. For example, the data on the website \cite{SpurCor} clearly demonstrates a strong correlation (87\%) between the age of Miss America and the number of murders by steam, hot vapors and hot objects. A goal in machine learning is to build algorithms that focus on the true causal relations and neglect such spurious correlations . 

Recently various interesting approaches have been developed towards identifying causal correlations in machine learning. A recurring idea in this context is to leverage data that is collected in different environments. Correlations that are universal or invariant over these different environments are more likely to be causal and an algorithm that specifically learns these relations is expected to be more robust towards new environments. An approach that is based on this concept is Invariant Risk Minimization (IRM) which estimates invariant correlations across multiple training distributions \cite{arjovsky2020invariant}. Another related approach revealed that a learner that is based on the correct causal structure adapts faster to (sparse) distributional changes \cite{bengio2019metatransfer, ke2019learning}.

\begin{figure}
    \includegraphics[width=0.7\columnwidth]{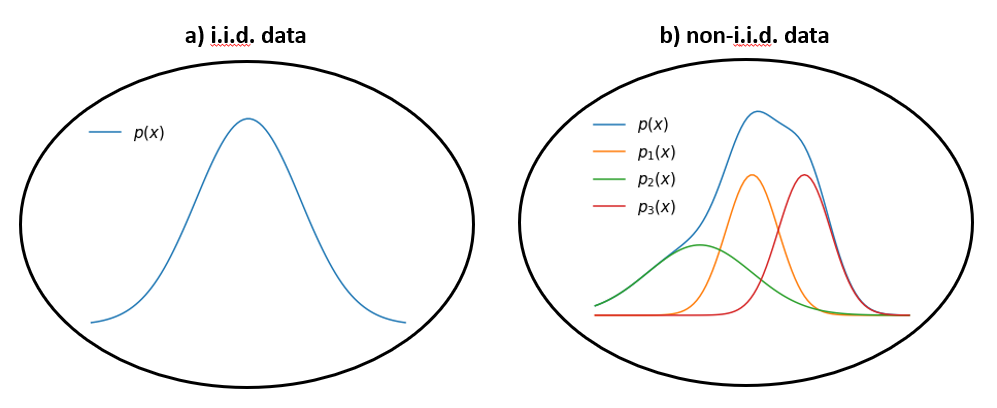}
    \caption{A schematic overview of the distribution $p(x)$ of (a) a data set with an i.i.d. distribution and (b) a data set with a non-i.i.d. distribution. In the latter case the data is composed of several subpopulations that could be related to a latent variable. }
    \label{fig_iid}
\end{figure}

In the present work we explore the possibility of leveraging data that is not i.i.d. towards more robust algorithms. The general idea is to divide the training data in different subpopulations and to introduce a regularization that penalizes correlations that are not invariant over these different subsets. An important advantage of this approach is that it is not necessary to collect data from different environments. The general concept is schematically presented in Fig. \ref{fig_iid} where (a) a distribution of i.i.d. data is presented and (b) of non-i.i.d. data. A similar behavior can be seen in the Simpson's paradox where a trend can be seen in several groups but disappears when the groups are combined \cite{simpson1951interpretation, pearl2016causal}. In case the different subgroups are not taken into account one arrives at faulty conclusions.

More specifically, this general idea is demonstrated in this paper with a classification task of images of cats and dogs. A data set is used that contains various different breeds which is considered as an unknown latent variable. Due to this the data is composed of different subpopulations and violates the i.i.d. property. Transfer learning is used to transform the images to numeric features with the ResNet-50 architecture \cite{He_2016_CVPR}. To obtain an out-of-distribution test set k-means clustering is used and one of the clusters is considered as the test set. The remaining data is used for training on which we again use k-means clustering to divide it in different subpopulations. A logistic regression is then fitted on each of these subsets and we use the full training set to fit a logistic regression with a regularization that penalizes weights that deviate from the weights learned on the subsets. This regularization favors correlations that are universal over the subpopulations. Finally, we compare the performance of our approach on the out-of-distribution test set to a more traditional $l2$-regularized logistic regression.




The paper is organised as follows: in the second section we discuss the data preprocessing steps. The third section introduces the new approach about how non-i.i.d. data can be leveraged in a regularization scheme towards more robust algorithms and how it is applied to the classification of cats and dogs with different breeds. The results are shown in section 4 and in section 5 the conclusions are presented together with an outlook to future work.

\section{Preprocessing}
\subsection{Images}
\begin{figure}
    \includegraphics[width=0.7\columnwidth]{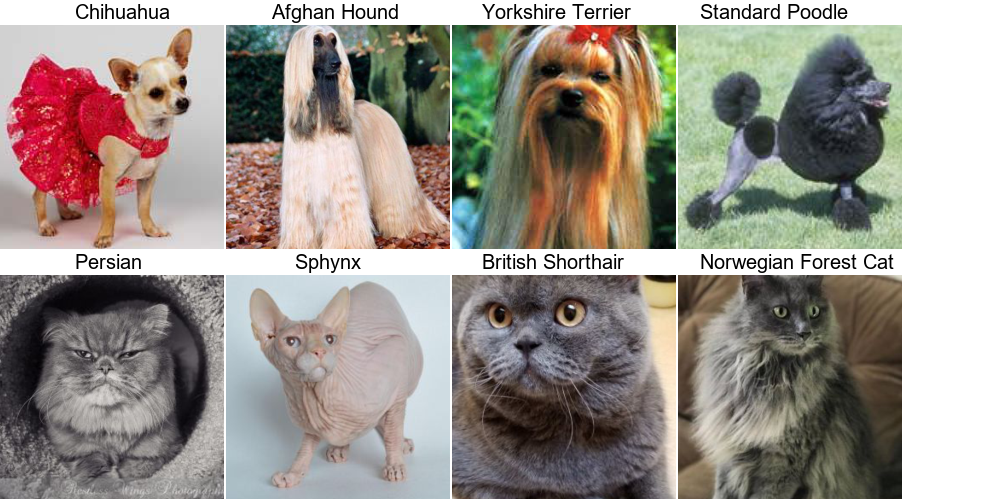}
    \caption{Some example images of the data set with the corresponding breed indicated. This shows that within the data set there are subpopulations that are determined by the latent variable breed.}
    \label{fig_cats_and_dogs}
\end{figure}

The considered task is the classification of images of dogs and cats. Since we are interested in data that is not i.i.d. a dataset is considered that contains various breeds. For the dog images the Stanford Dogs data set is used which consists of 20.580 pictures that are roughly evenly distributed over 120 breeds of dogs \cite{KhoslaYaoJayadevaprakashFeiFei_FGVC2011}. For the cat images a dataset is used that is shared on Kaggle and contains 125k images distributed over 67 breeds \cite{Kaggle_cats} (for this data set the distribution over the breeds is highly skewed and the labeling of the breeds is not so accurate). To avoid issues with an unbalanced dataset we remove the images denoted as domestic cats and then take a random subsample of 1/4th for the cat dataset resulting in 22.844 cat pictures. In Fig. \ref{fig_cats_and_dogs} some illustrative samples are presented which clearly reveal the different breeds corresponding to different properties and subpopulations.  

\subsection{transfer learning with ResNet-50}
\begin{figure}
    \includegraphics[width=0.7\columnwidth]{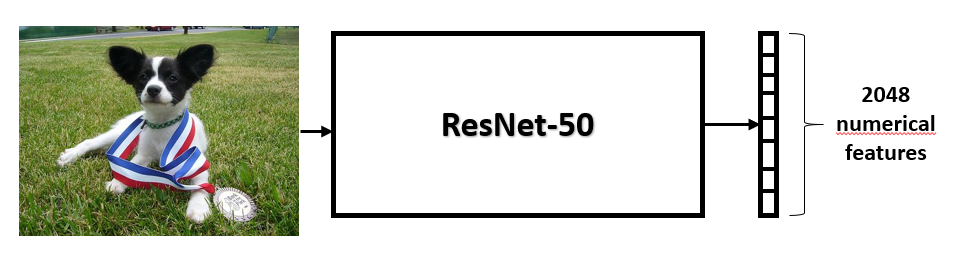}
    \caption{Schematic overview of the transfer learning approach that is used to transform the images to numerical features. The images are fed to the RESNET-50 architecture from which the final fully connected layer is removed resulting in 2048 features.}
    \label{fig2}
\end{figure}

To transform the images to numeric features we use transfer learning with ResNet-50 \cite{He_2016_CVPR}. The final fully connected layer is removed such that we are left with 2048 numerical features (see Fig. \ref{fig2}). Before feeding the images to ResNet-50 they are resized to 256 pixels and a center crop of 224 $\times$ 224 pixels is taken. After that the pixel values are normalized for each RGB channel with a standard scaler (with the means and standard deviations recommended for the ResNet model). In the following the resulting numerical vector for a sample $i$ with 2048 numeric features is denoted as ${\vec{x}_i}$.

\subsection{Out-of-distribution test set \label{OOD}}
Instead of the traditional procedure where a random subsample is taken from the data set as a test set we are interested in an out-of-distribution test set with a covariate shift. Since the labeling of the cat images is not very accurate and it is at first sight not always clear which breeds are closely related we use k-mean clustering to determine which samples are closely related and which are not. Both the cat and the dog samples are separately divided in 5 clusters (see table \ref{table:1} for the number of samples in the clusters) and the test set is composed by combining a cluster of the cat samples with a cluster of the dog samples. 

\begin{table}[h!]
\centering
\begin{tabular}{ |c|c c c| } 
 \hline
 \textbf{cluster }& $n_{cats}$ & $n_{dogs}$ & $n_{dogs}/n_{tot}$ \\ 
 \hline
 \textbf{1}  & 3875 & 3145 & 45\%\\ 
 \textbf{2} & 6022 & 5394 & 47\%\\ 
 \textbf{3} & 5042 & 4319 & 46\%\\ 
 \textbf{4} & 5928 & 3672 & 38\%\\ 
 \textbf{5} & 1977 & 4050 & 67\%\\ 
 \hline
\end{tabular}
\caption{The different clusters with the corresponding numbers of cat and dog samples. The final column gives the percentage of dog samples in each cluster. The cats and dogs clusters are then combined and the first cluster is withheld as test set.}
\label{table:1}
\end{table}

\begin{figure}
    \includegraphics[width=0.7\columnwidth]{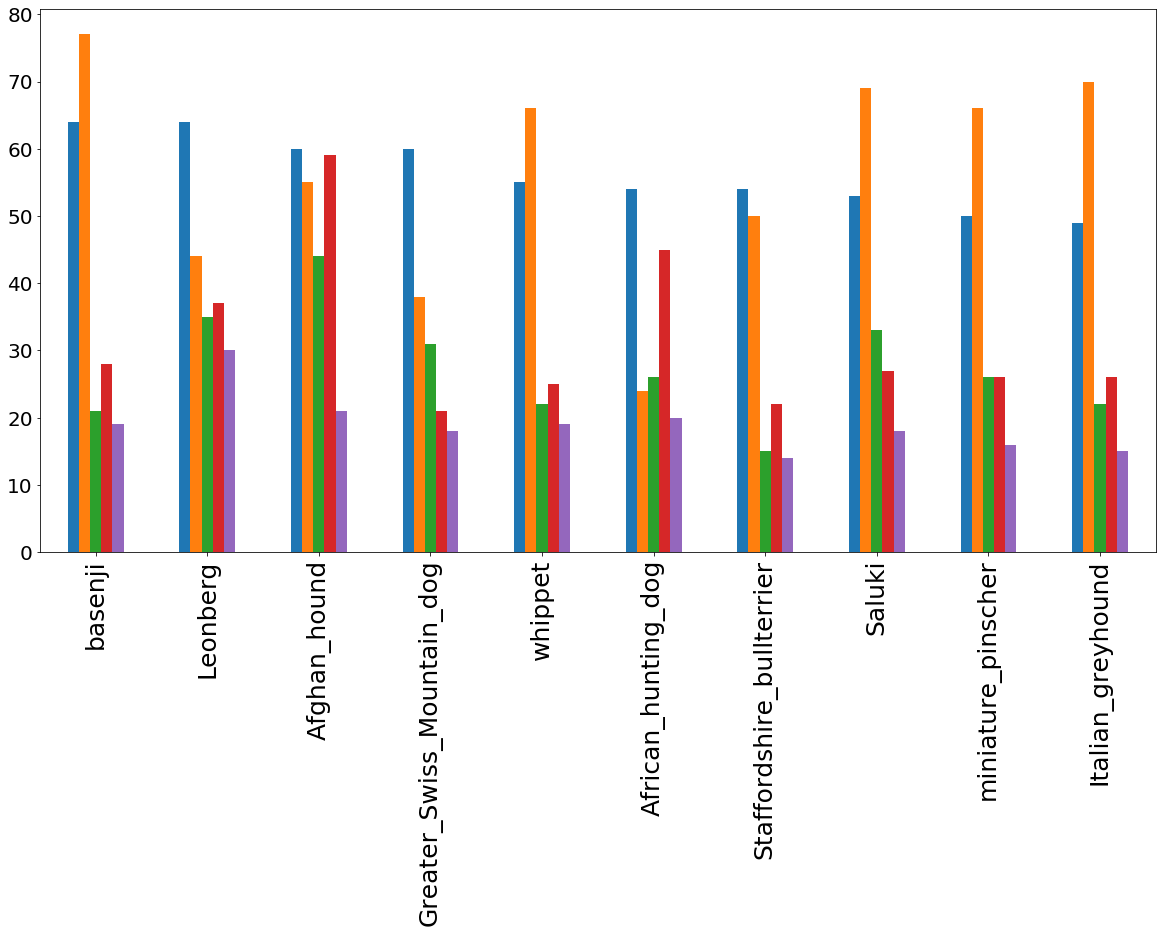}
    \caption{The distribution of the cluster samples over the breeds for the dog images. The breeds are ordered in descending contribution to cluster $1$ (blue) and for clarity only 10 breeds are presented. This shows that the cluster distributions are not uniform over the breeds and the clusters correspond to different distributions. }
    \label{fig_dog_clus}
\end{figure}

In Fig. \ref{fig_dog_clus} the distribution over the breeds is presented for the different clusters with dog samples (for clarity only 10 out of the 120 breeds are presented). This clearly shows that the cluster distribution in not uniform over the breeds and the clusters correspond to different distributions. 

\section{Non-i.i.d. regularization with logistic regression}

\subsection{logistic regression with $l_2$ regularization \label{logreg}} 
We start by giving a short overview of the basic features of logistic regression with $l_2$-regularization since a similar approach will be used (for more details on logistic regression the interested reader is referred to standard textbooks on machine learning such as \cite{bishop}). Logistic regression allows to determine a mapping from image features to the class probabilities. To avoid possible issues with unstable weights related to correlated features the input features $\vec{x}$ are first transformed to the linearly independent principal components $\vec{PC}$ (this transformation is determined form the training set). The principal assumption of logistic regression is that the probability $P(\text{dog}|\vec{x}_i)$ that a sample $i$ with features ${\vec{x}_i}$ (and principal components $\vec{PC_i}$) corresponds to the dog class can be written as a sigmoid function:
\begin{equation}
P(\text{dog}|\vec{x}_i) = \frac{1}{1+\exp[-w_0 -  \vec{w}.\vec{PC}_i]}.
\end{equation}
Where $\vec{w}$ is the vector with the weights that are together with the bias term $w_0$ determined with maximum likelihood estimation. This corresponds to minimizing the negative logarithm of the likelihood (corresponding to the cross-entropy error function):
\begin{align}
\mathcal{L} = - &\sum_i\{y_i\log[P(\text{dog}|\vec{x}_i)] + (1-y_i)\log[1 - P(\text{dog}|\vec{x}_i)]\} \nonumber \\
				& + \mathcal{R}(\vec{w})
\end{align}
where $y_i$ is the label of sample $i$ (1 if it is a dog and 0 if it is a cat) and $\mathcal{R}(\vec{w})$ is a regularization term. The purpose of this term is to reduce overfitting by imposing constraints on the weights. A common choice is $l_2$ regularization corresponding to the following regularization term:
\begin{equation}
\mathcal{R}_{l_2}(\vec{w}) = \lambda ||\vec{w}||^2,
\end{equation}
where the $l_2$-norm is considered and $\lambda$ is a hyperparameter. This parameter can for example be determined by cross validation. The training set is then randomly split in a number of disjoint sets and the weights are fitted to all but one set (the validation or hold-out set). The generalization error is then estimated by calculating the performance on the validation set. The optimal value for the hyperparameter is determined as the one corresponding to the lowest generalization error.

\subsection{Non-i.i.d. regularization}
As discussed in the introduction a recurring idea in approaches towards more robust algorithms is to leverage data that is collected in different environments. Here we instead leverage the diversity of the training set which is composed of data from different subpopulations. In the considered case this is clear as the different breeds can be seen as subpopulations. To identify the subpopulations the same procedure is used as in section \ref{OOD} to obtain the out-of-distribution test set. The k-means clustering algorithm is used to split the training set in clusters $c$ of similar data ($\mathcal{C}$ is used to denote the set of clusters). Then we leverage these clusters by introducing the following non-i.i.d. (ni) regularization term :
\begin{equation}
\mathcal{R}_{ni}(\vec{w}) = \alpha\sum_{c\in\mathcal{C}}||\vec{w} - \vec{w}_c||^2,
\label{eq:1}
\end{equation}
where $\alpha$ is a hyperparameter and $\vec{w}_c$ are the weights obtained from a logistic regression on the samples in cluster $c$. This regularization favors correlations that are universal over the different clusters by penalizing weights that are different from the weights obtained for the separate clusters. Weights corresponding to relations that are also present in the separate clusters are thus encouraged.

To determine the hyperparameter $\alpha$ a procedure similar to cross validation (discussed in section \ref{logreg}) is used. The difference is that the splitting of the training data is not random. Instead, the training data is split in clusters with the k-means clustering algorithm as discussed before. The logistic regression is then fitted to all data (with the ni-regularization from Eq. (\ref{eq:1}))except one cluster which is used to calculate the generalization error. Note that the same clusters $\mathcal{C}$ from Eq. (\ref{eq:1}) are used for the cross validation. A final estimate is obtained by repeating this procedure over all clusters as holdout set and averaging the generalization error.

\section{Results}
\begin{figure}
    \includegraphics[width=0.7\columnwidth]{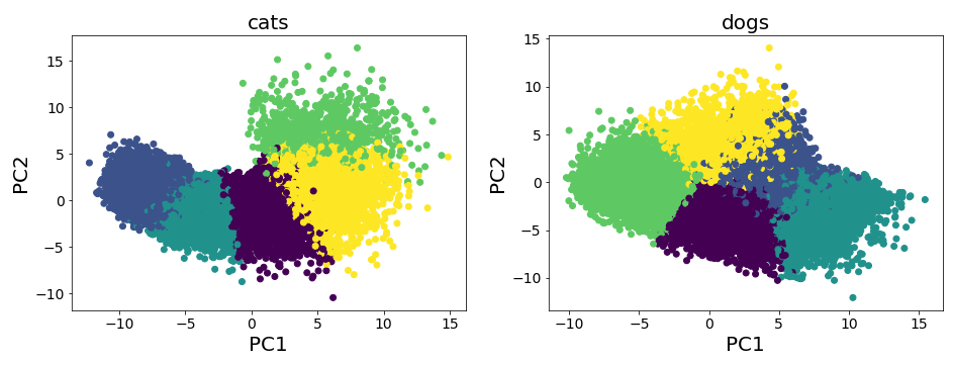}
    \caption{The training set is divided in 5 clusters with the k-means algorithm for the cats (left figure) and the dogs (right figure). The colors indicate the clusters and the values corresponds to the two main principal components (PC1 and PC2).}
    \label{fig_clus}
\end{figure}

First, k-means clustering is used to split the training set in 5 sets separately for the dog and cat samples (see Fg. \ref{fig_clus}). These clusters are also used for the cross validation procedure to determine the hyperparameter $\alpha$ in Eq. (\ref{eq:1}).

\begin{figure}
    \includegraphics[width=0.7\columnwidth]{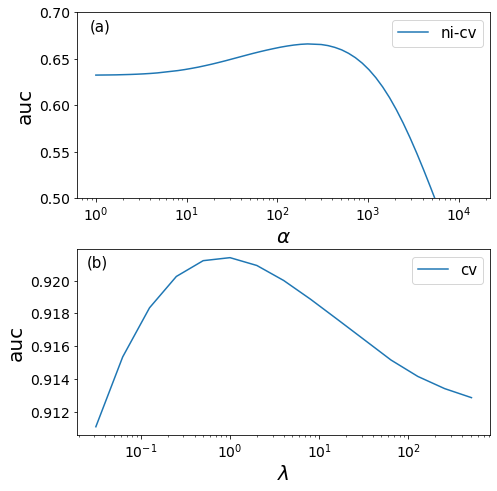}
    \caption{The area under the ROC-curve (auc) as a function of (a) $\alpha$ for the non-i.i.d. (ni) regularisation and of $\lambda$ (b) for $l_2$-regularisation. The optimal values for the hyperparameters correspond to the values where the auc is maximal.}
    \label{fig_cv}
\end{figure}

For comparison a logistic regression with $l_2$-regularisation is also fitted where the parameter $\lambda$ is determined by the usual cross validation (discussed in section \ref{logreg}). To determine and compare the performance betwee the appoaches we consider the ROC-curve and the area under the ROC-curve (auc). In Fig. \ref{fig_cv} the auc is presented as a function of (a) $\alpha$ for the non-i.i.d. (ni) regularization and (b) $\lambda$ for the usual cross validation (cv). This allows to determine the optimal value for these hyperparameters corresponding to the highest performance in terms of the auc on the hold-out set. 

It is clear from Fig. \ref{fig_cv} that we obtain a larger auc value for the $l_2$-regularization with respect to the ni-regularization. This could be expected since for the the $l_2$-regularization the hold out set is chosen randomly and has the same distribution as the rest of the training set. For the ni-regularization on the other hand the validation set is chosen explicitly to not have the same distribution as the training set making it harder to obtain a good performance. The final goal is to have a robust performance on data from a new environments and to appropriately compare the two approaches we have to consider the performance on the same data set. 

\begin{figure}
    \includegraphics[width=0.7\columnwidth]{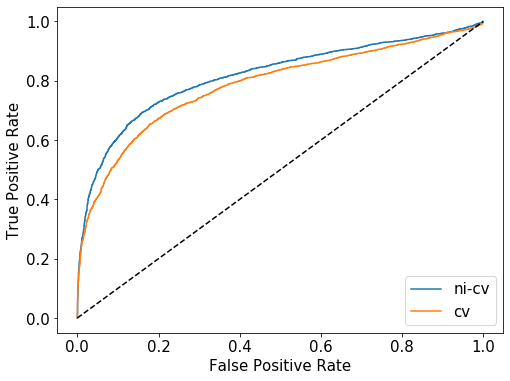}
    \caption{The ROC-curve on the out-of-distribution test set obtained with the non-i.i.d. (ni-cv) regularization and with $l_2$-regularization (cv). The corresponding values for the area under the curve (auc) are 0.817 for ni-regularization and 0.785 for $l_2$-regulariation. This reveals that we obtain a better perfomance of the logistic regression on the out-of-distribution test set with the ni-regularization.}
    \label{fig_roc}
\end{figure}

This is done in Fig. \ref{fig_roc} where the ROC-curve is presented on the same out-of-distribution test set (as discussed in section \ref{OOD}) with the hyperparameters corresponding to the optimal values in Fig. \ref{fig_cv}. This reveals that the ni-regularization performs better as the corresponding ROC-curve is strictly above the one obtained with $l_2$-regularization. The resulting auc on the test set is 0.817 with the ni-regularization and 0.785 with the $l_2$-regularization. 

\section{Conclusions and Outlook}
We have presented a new approach towards robust machine learning that has the advantage that only a single training set is needed and no data from different environments has to be collected. Instead the approach leverages the fact that many datasets are not i.i.d. and different subpopulations are present within a single data set. Once these subsets are found a regularization scheme is used to learn relations that are invariant over the different subpopulations. Intuitively these relations are more likely to be causal instead of nonsensical spurious correlations. 

This approach is demonstrated on a classification task of cats and dogs by introducing a non-i.i.d. regularization term that penalises weights that are not universal over different clusters in the training data. This results in a more robust algorithm with a better performance on an out-of-distribution test set as compared to a more traditional $l_2$-regularisation. 

This analysis is presented for a (random) single choice of test set and clusters. Our results reveal that it is possible to have a more robust algorithm with the proposed ni-regularization. A more in depth analysis of the robustness of these results with respect to different test sets and clusters is postponed for future work. The main conclusion is that it is possible to obtain a better performance on an out-of-distribution test set with ni-regularization.

These results could pave the way towards more robust machine learning algorithms in general. For now a specific classification task is considered but it is expected that the general concepts are applicable much more broadly. The only requisite is that the dataset on which the algorithm is trained is not i.i.d. which is independent of the specific application or algorithm architecture. 

Many of the separate components that have been introduced can be further optimised. It would for example be interesting to examine how other clustering algorithms or other unsupervised techniques could improve the identification of different subpopulations in the training set. Other interesting future work would be to integrate the different steps into an end-to-end training with a loss function that concurrently separates the training data in different subpopulations and learns patterns that are invariant over these subsets.

\bibliographystyle{plain}
\bibliography{refs}

\end{document}